\documentclass{bmvc2k}
\usepackage{multirow}
\usepackage{tikz}
\usepackage{microtype}
\usepackage[symbol]{footmisc}
\usepackage{floatrow}
\newfloatcommand{capbtabbox}{table}[][\FBwidth]

\usetikzlibrary{matrix,shapes,arrows,positioning,chains}

\title{Deep Metric Learning for Ground Images}

\addauthorw{Raaghav Radhakrishnan}{raaghavradhakrishnan@gmail.com}{ 1}{2}
\addauthorw{Jan Fabian Schmid}{SchmidJanFabian@gmail.com}{ 1}{3}
\addauthor{Randolf Scholz}{scholz@ismll.uni-hildesheim.de}{ 2}
\addauthor{Lars Schmidt-Thieme}{schmidt-thieme@ismll.uni-hildesheim.de}{ 2}
\addinstitution{
    Robert Bosch GmbH\\
	Hildesheim, Germany
}
\addinstitution{
    University of Hildesheim\\
	Hildesheim, Germany
}
\addinstitution{
    Goethe University\\
	Frankfurt am Main, Germany
}
\runninghead{\footnotesize Radhakrishnan, Schmid, Scholz, Schmidt-Thieme}{\footnotesize Ground Image Retrieval}

\def\etal{\emph{et al}\bmvaOneDot}

\newcommand{\Variable}[1]{\mathrm{\MakeLowercase{{\textit{#1}}}}}

\newcommand{\Set}[1]{{\mathrm{\MakeUppercase{\mathcal{{#1}}}}}}

\newcommand{\vd}[0]{{\Variable{d}}}
\newcommand{\ve}[0]{{\Variable{e}}}

\newcommand{\vk}[0]{{\Variable{k}}}

\newcommand{\vn}[0]{{\Variable{n}}}
\newcommand{\vo}[0]{{\Variable{o}}}
\newcommand{\vp}[0]{{\Variable{p}}}
\newcommand{\vq}[0]{{\Variable{q}}}
\newcommand{\vr}[0]{{\Variable{r}}}

\newcommand{\vx}[0]{{\Variable{x}}}

\newcommand{\sQ}[0]{{\Set{Q}}}
\newcommand{\sR}[0]{{\Set{R}}}

\newcommand{\RatK}[0]{\mathrm{R@}\vk}
\newcommand{\RatKi}[1]{\mathrm{R@}#1}
\newcommand{\RiatK}[1]{\mathrm{R\textsubscript{$#1$}@}\vk}
\newcommand{\RiatKj}[2]{\mathrm{R\textsubscript{$#1$}@}#2}

\begin{document}

\renewcommand{\thefootnote}{\fnsymbol{footnote}}
\maketitle
\footnotetext[1]{Equal contribution.}
\renewcommand*{\thefootnote}{\arabic{footnote}}

\begin{abstract}
Ground texture based localization methods are potential prospects for low-cost,
high-accuracy self-localization solutions for robots.
These methods estimate the pose of a given query image,
i.e. the current observation of the ground from a downward-facing camera,
in respect to a set of reference images whose poses are known in the application area.
In this work,
we deal with the initial localization task,
in which we have no prior knowledge about the current robot positioning.
In this situation,
the localization method would have to consider all available reference images.
However,
in order to reduce computational effort and the risk of receiving a wrong result,
we would like to consider only those reference images that are actually overlapping with the query image.
For this purpose,
we propose a deep metric learning approach that retrieves the most similar reference images to the query image.
In contrast to existing approaches to image retrieval for ground images,
our approach achieves significantly better recall performance
and improves the localization performance of a state-of-the-art ground texture based localization method.
\end{abstract}

\section{Introduction}
\begin{figure}[t]
	\includegraphics[width=0.16\textwidth]{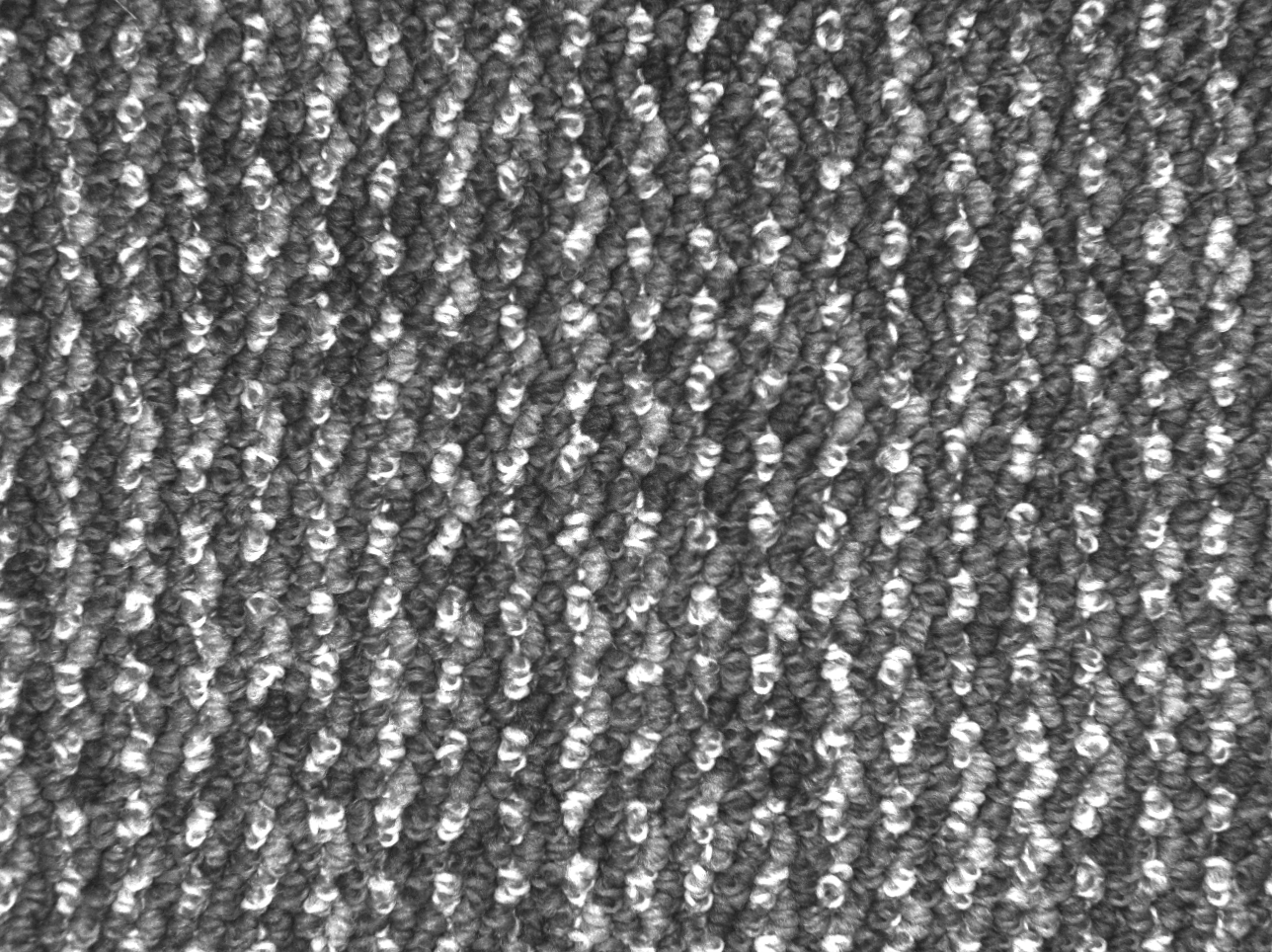}
	\includegraphics[width=0.16\textwidth]{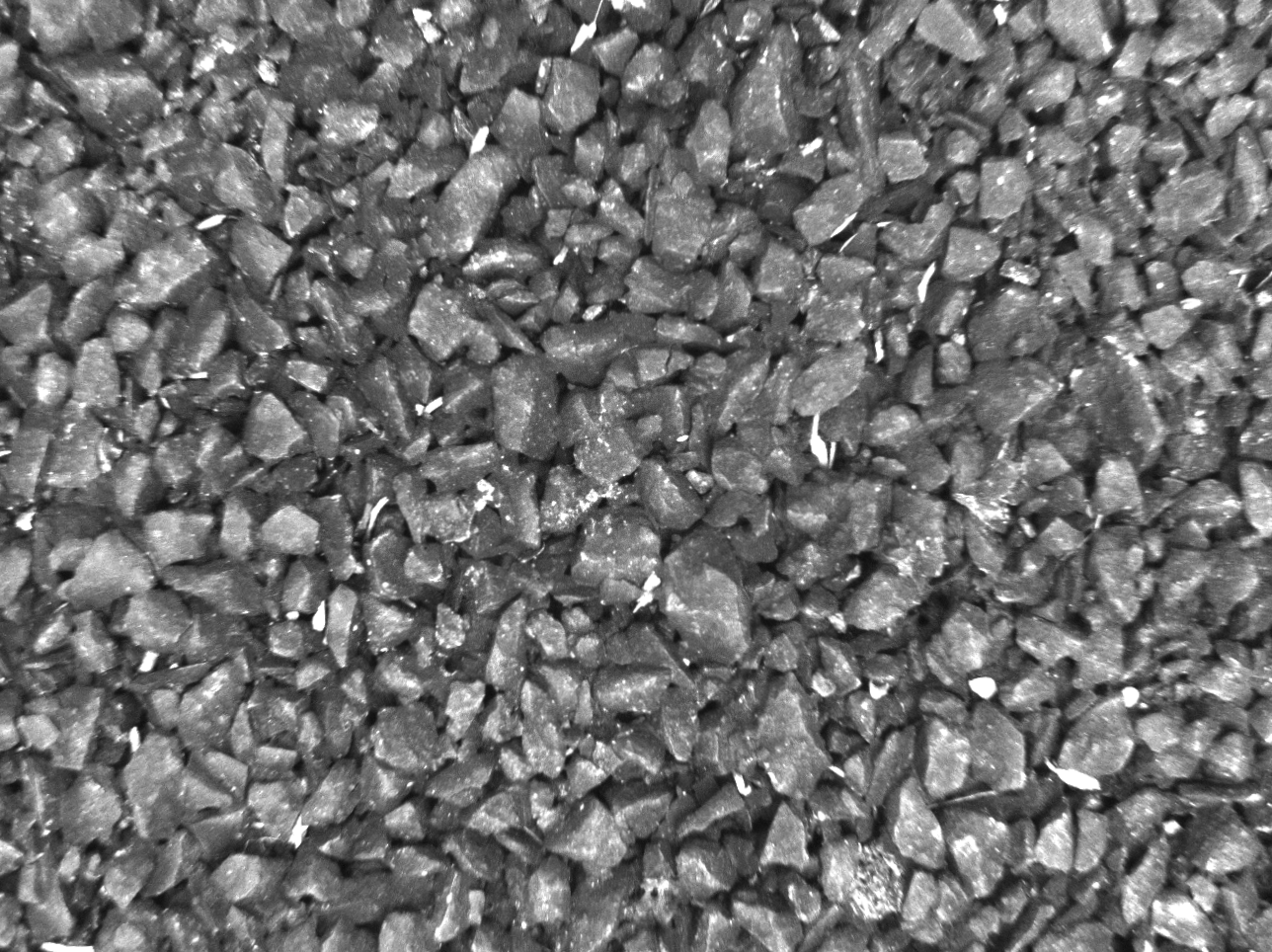}
	\includegraphics[width=0.16\textwidth]{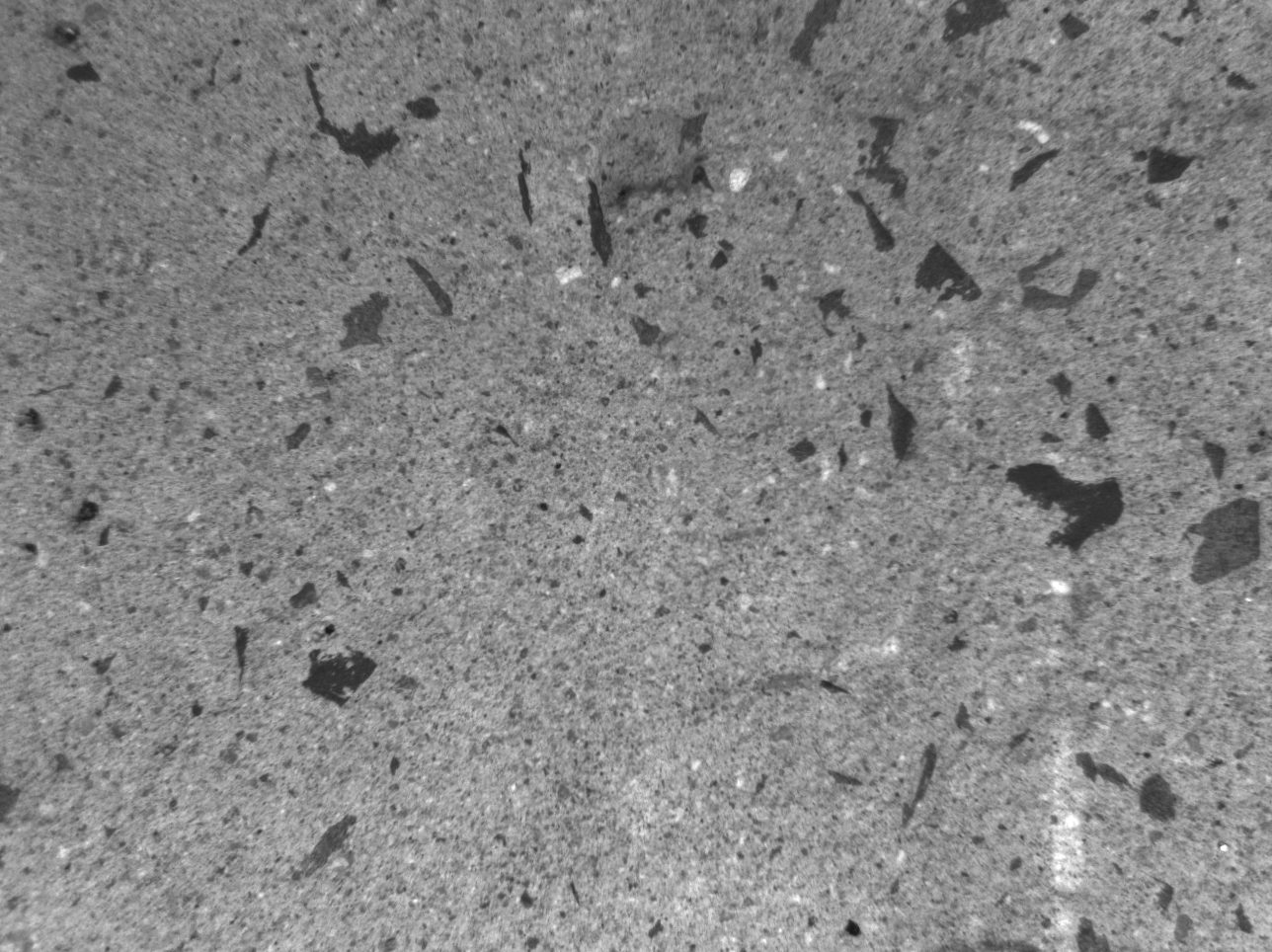}
	\includegraphics[width=0.16\textwidth]{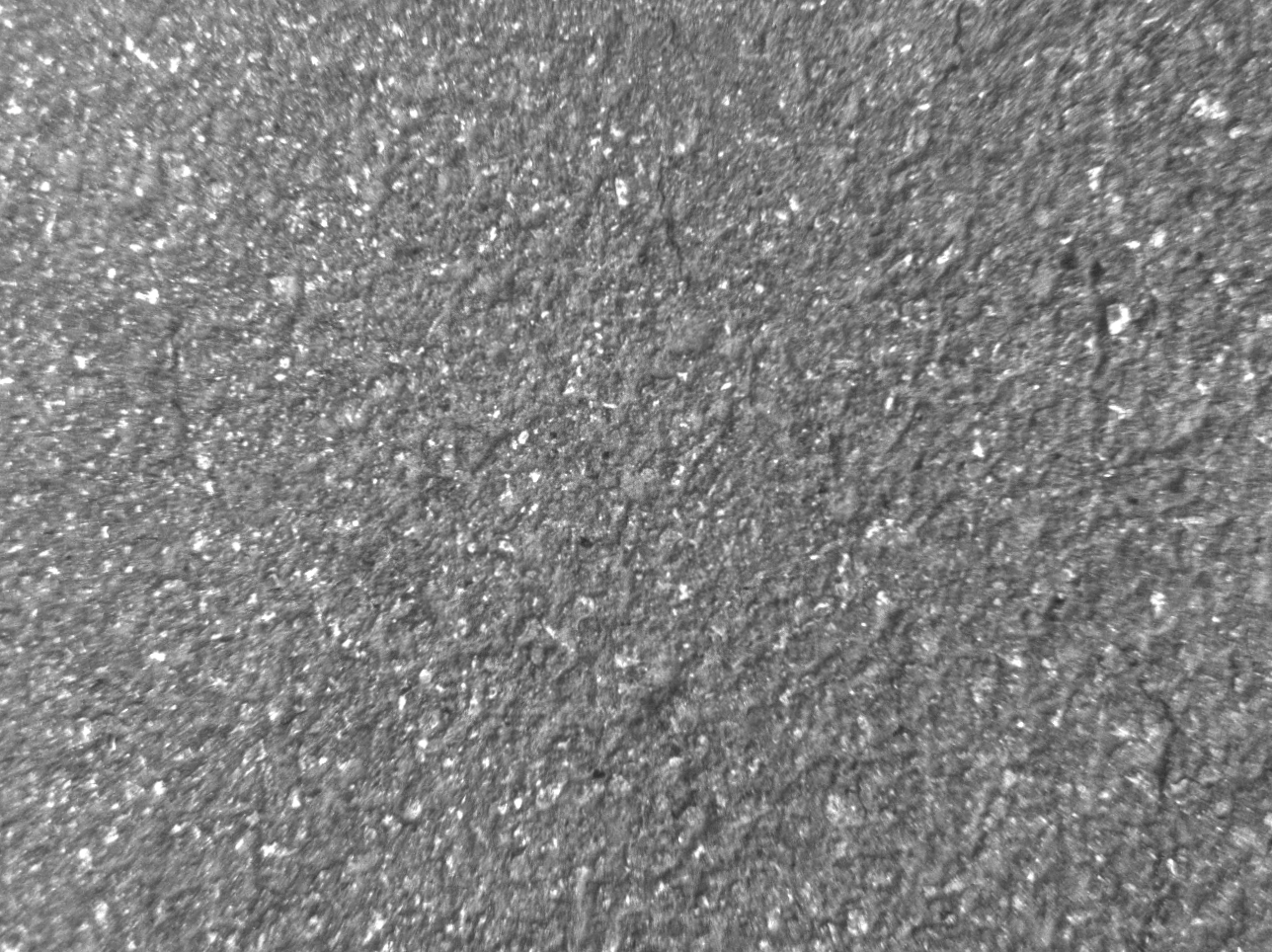}
	\includegraphics[width=0.16\textwidth]{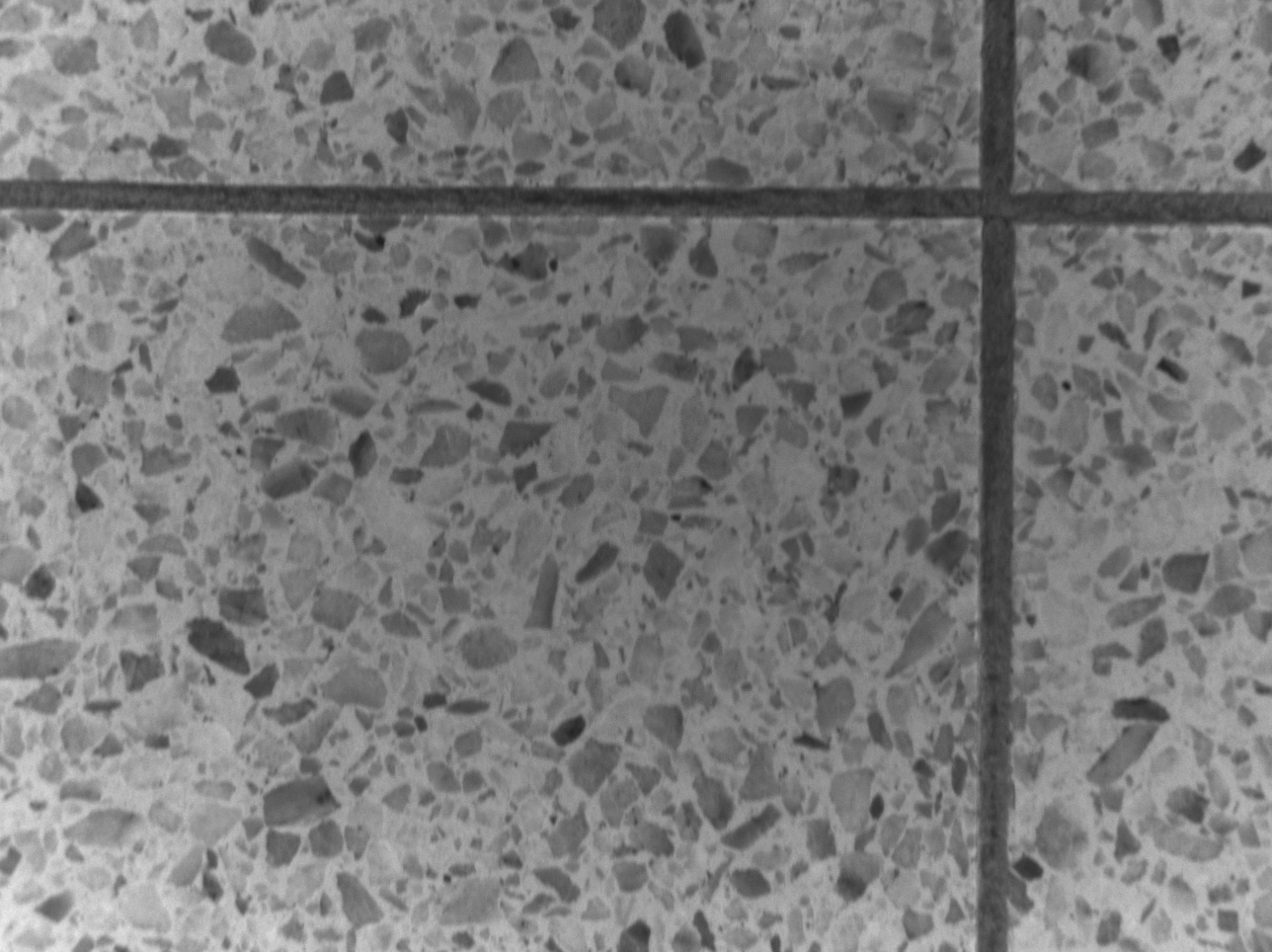}
	\includegraphics[width=0.16\textwidth]{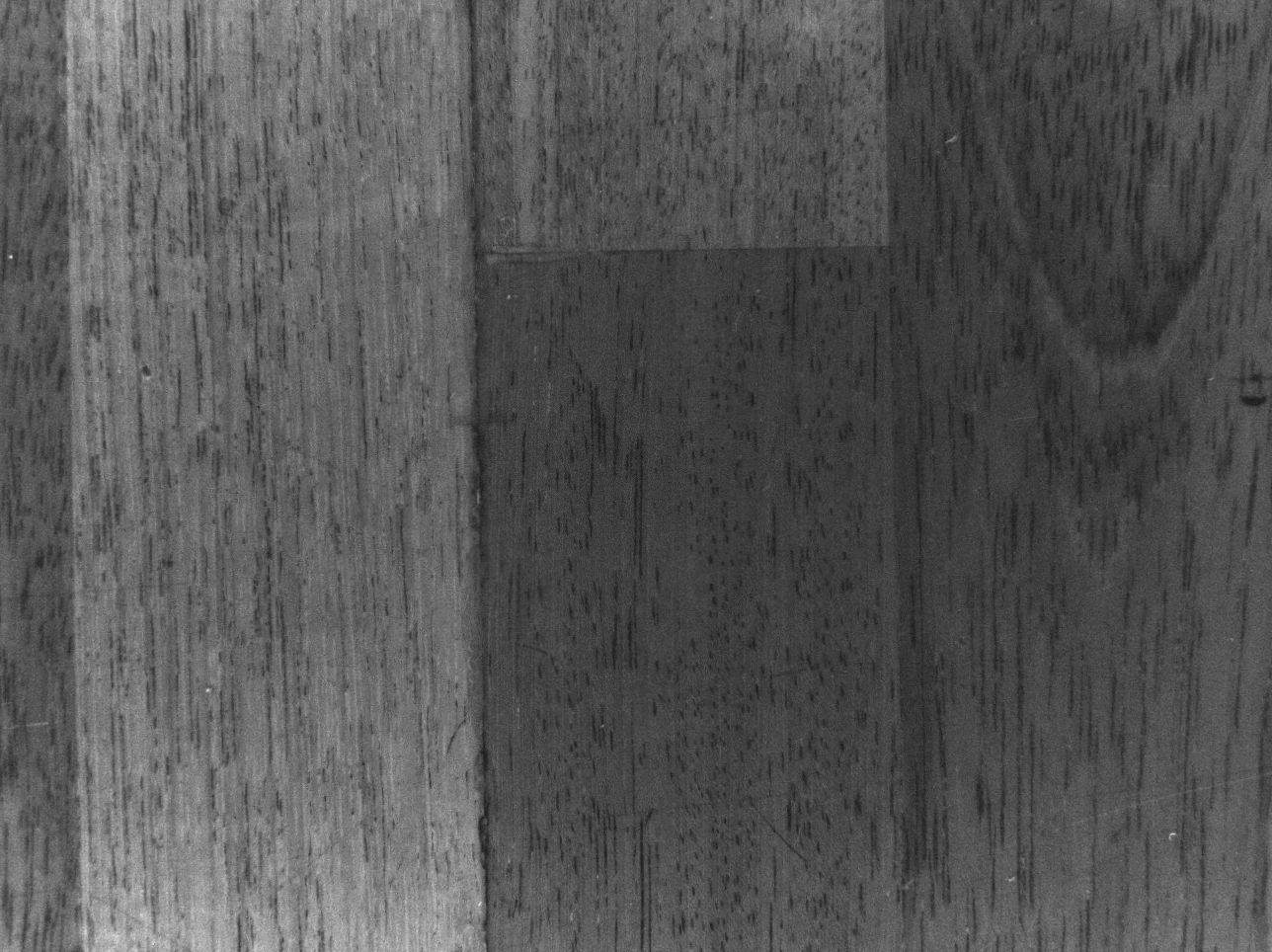}
   \vspace{-0.15cm}	
   \caption{Examples of the examined ground textures from the image database of  Zhang~\etal~\cite{zhang2019high}.
   From left to right: carpet, coarse asphalt, concrete, fine asphalt, tiles, and wood.}
\label{fig:textures}
\end{figure}

High-accuracy localization capabilities are required to perform tasks like freight
and passenger transport autonomously~\cite{schmid2020ground}.
A sound approach to this task is to localize visually based on unique feature constellations on the ground,
as it enables localization accurate to a few millimeters in indoor and outdoor scenarios~\cite{zhang2019high}
without requiring anything else but the natural floor covering.
These methods work even if the surrounding is occluded
and they can be made independent of external lighting conditions
if the robot is equipped with its own ground illumination.
Ground textures such as carpet, asphalt,
and concrete may look indistinguishable to the human eye (see Figure \ref{fig:textures}),
but,
they contain unique arrangements of visual features
allowing for unambiguous identification of a specific spot on the ground~\cite{schmid2020rndKPs}.

Having overlapping images available,
state-of-the-art methods rely on feature-based localization,
e.g.~\cite{zhang2019high, chen2018streetmap, schmid2020ground, schmid2020rndKPs, Kozak_Ranger}.
First,
they detect local visual features,
such as blobs with SIFT~\cite{lowe2004distinctive},
which is well suited for ground images~\cite{schmid2019survey}.
Here,
each detected feature has two properties:
a keypoint defining the location of its corresponding image patch and a descriptor representing
its visual content.
Then,
the extracted features are matched with each other to identify correspondences,
which are used to estimate the pose,
i.e. position and orientation,
of one image (the query image) in respect to the other images (the reference images)~\cite{zhang2019high}.
\begin{figure}[t]
   \includegraphics[width=0.999\textwidth]{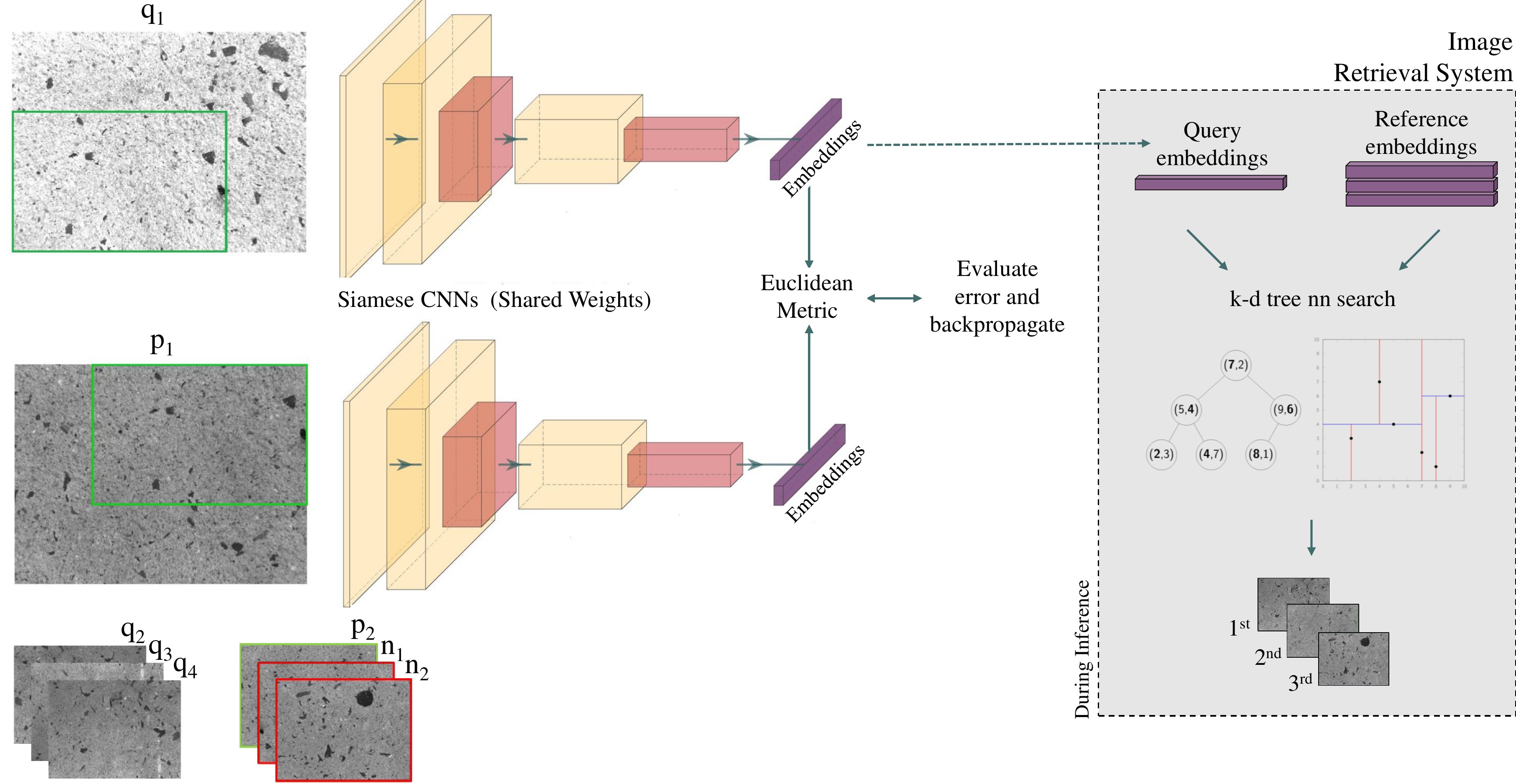}
   \vspace{-0.15cm}
   \caption{
    Overview of the proposed image retrieval approach.
    We use a Siamese CNN architecture, 
    whose final layer activations represent image embeddings.
    During training,
    the model tries to predict the overlap between randomly sampled pairs of images.
    At inference,
    we use a k-d tree to compare the embedding of the query image
    against a database of the learned embeddings of the reference images to find the closest matches.
    }%
\label{fig:sn_nn}
\end{figure}

In this work,
we consider the task of map-based localization,
i.e. a map consisting of a set of reference images with known poses is available,
and,
in the map coordinate system,
we want to estimate the pose of an independently recorded query image.
A 
challenging manifestation of this task is the initial localization,
which is necessary when we have no knowledge about the current location of the robot,
e.g. after restart or in recovery mode after mislocalization.
This task is difficult,
because it does not allow to restrict the search space for the current query image pose
based on an existing approximate pose estimate.
Therefore,
features of all available reference images have to be considered in the feature matching step,
increasing the computational effort,
increasing the number of incorrectly proposed feature correspondences,
and therefore,
increasing the chance of mislocalization~\cite{schmid2020rndKPs}.
A possible solution 
was introduced by Chen~\etal~\cite{chen2018streetmap}.
With the idea in mind that they would like to consider only those reference images
that are actually overlapping with the query image,
they propose to apply a Bag-of-Words (BoW) image retrieval approach
to obtain the most similar reference images to a given query image.
Then,
only the features of the retrieved most-similar reference images are used for the subsequent feature matching
and pose estimation steps.
The solution of Chen~\etal makes use of the hand-crafted SURF~\cite{Bay_SURF} feature extraction method.
However,
methods based on deep learning have begun to outperform hand-crafted methods in many computer vision areas%
~\cite{krizhevsky2012imagenet,han2015matchnet,zagoruyko2015learning}. 
Correspondingly,
we propose a deep learning approach to image retrieval of ground images,
based on deep metric learning,
to substitute the use of hand-crafted feature extractors and the BoW technique.
Still,
BoW is the current state of the art for the retrieval of ground images,
which was shown to achieve good performance~\cite{chen2018streetmap, schmid2020ground}.
This is why we perform an in-depth evaluation of this approach,
searching for optimal parametrization to examine the method in its best possible configuration.

Our method learns similarities between ground images to represent them as compact embeddings,
i.e. image descriptors,
for image retrieval.
It consists of a Convolutional Neural Network (CNN) that is trained in Siamese fashion,
using a loss function that is adopted from S{\'a}nchez-Belenguer~\etal~\cite{sanchez2020rise}.
Subsequently,
we employ a k-d tree to find the reference images with most similar embeddings to that of the query image.
Our results show that our method outperforms BoW image retrieval,
with significantly higher recall values especially for the most difficult cases.
Also,
we employ the localization pipeline of Schmid~\etal~\cite{schmid2020ground}
to estimate the query image pose with the retrieved images.
Here,
our method again outperforms the BoW approach,
and slightly outperforms initial localization without image retrieval.

This paper contributes a deep metric learning approach
to represent ground images with compact descriptors.
It is the first method that can be trained specifically for image retrieval of ground images.
We introduce an evaluation framework for the task,
and,
for BoW,
the current state-of-the-art approach to ground image retrieval,
we investigate proper design decisions,
such as the choice of the detector-descriptor pair.
Still,
we show that our method clearly outperforms BoW
both in image retrieval recall and the resulting localization performance.

\section{Related work}
To the best of our knowledge,
we propose the first deep learning approach that can be applied directly to ground image retrieval.
BoW,
the current state of the art for this task, is based on the aggregation of manually designed descriptors.
Existing deep learning approaches to image retrieval are not applicable to the task.
This is, for example, because they explicitly learn camera poses,
like PoseNet~\cite{kendall2017geometric},
which prevents an application for our problem where training and testing images are from separate places,
or, as~\cite{AP-GeM}, \cite{TL-MAC}, \cite{DELF}, and~\cite{R-MAC},
they are trained with a classification loss,
having places correspond to classes.
Such a classification of images is not directly applicable to our task,
as every query image would form its own class with an image-specific set of overlapping reference images.
In the following,
we present existing work on image retrieval for localization tasks and ground texture based localization.

\subsection{Image retrieval for place recognition and localization}
Several methods,
such as StreetMap~\cite{chen2018streetmap}
and RelocNet~\cite{balntas2018relocnet},
exploit a coarse-to-fine paradigm for localization.
They retrieve similar reference images to the query image globally,
followed by a fine-grained adjustment of the estimated query pose.
RelocNet~\cite{balntas2018relocnet} trains a CNN in Siamese fashion
to learn continuous camera pose descriptors with a metric loss based on camera frustum overlaps.
For localization,
RelocNet finds the nearest neighbor to the query image and uses it for pose estimation with a differential pose network.

Gordo~\etal~\cite{TL-MAC} developed a deep metric learning approach for image retrieval,
consisting of a CNN trained in Siamese-fashion with triplet ranking loss.
The authors employ a R-MAC pooling layer
that corresponds to a differentiable variant of the R-MAC~\cite{R-MAC} descriptor.
The triplets generated for training consist of an anchor query image,
a positive sample of an image from the same class as the query image,
and a negative sample of an image from another class.
The network learns to generate query image descriptors
that are more similar to that of the positive sample
than to that of the negative sample.
In the following,
we call this method TL-MAC,
which stands for triplet loss with R-MAC descriptor.

Revaud~\etal~\cite{AP-GeM} adapt TL-MAC.
Instead of using a triplet ranking loss,
they directly optimize for the mean Average Precision (AP),
considering large numbers of images at each training step.
Also,
they substitute the R-MAC pooling layer of TL-MAC with a Generalized-Mean (GeM) pooling layer.
We call this method AP-GeM.

S{\'a}nchez-Belenguer~\etal~\cite{sanchez2020rise} developed RISE,
an image retrieval based indoor place recogniser,
which we are building upon.
They create a 3D map of the environment with a laser and calibrated spherical camera mounted on a backpack.
For image retrieval,
they train a CNN in Siamese fashion using overlap information of image pairs.
The map is voxelized,
which is used to compute the content overlap of any two images:
for each image corresponding depth information is available,
therefore,
the set of visible mapped-voxels can be identified,
and the overlap is then computed as the number of common visible voxels.  
During training,
the network learns to predict image pair overlaps.
The activation of the final layers represent the image embeddings.
Then,
the network is optimized to minimize the error between the predicted dissimilitude (L2-Norm) of the images
and their actual 3D overlap.
Once the network is optimized,
an offline database is created from the reference image embeddings and their associated poses.
For place recognition,
query and reference embeddings are compared online using a k-d tree to retrieve overlapping reference images.

\subsection{Ground texture based localization methods}
We consider methods for map-based initial localization using only ground images.

Chen~\etal~\cite{chen2018streetmap} developed the StreetMap framework.
StreetMap extracts SURF~\cite{Bay_SURF} features.
Every query image feature is matched with its nearest neighbor
among the features extracted from the reference images.
The first-to-second ratio test~\cite{lowe2004distinctive} is applied for outlier rejection,
and,
the query image pose is estimated in a RANSAC fashion.
The authors propose a variant of this framework 
for initial localization.
It relies on Bag-of-Words (BoW) image retrieval to obtain the most similar reference images to the query image.
BoW is an aggregated descriptor~\cite{Duan_WeightedHashing}.
The first step for BoW is to build a visual vocabulary.
This means that local visual features are extracted from a set of training images,
and subsequently clustered into groups of similar features.
Each cluster represents a visual word.
Then,
in order to compute a BoW representation of an image,
the visual vocabulary is used to map local visual features from an image to their corresponding visual words,
based on their descriptor values.
This mapping allows to quantize continuous feature descriptors
and the 
visual words have lower dimensionality than the feature descriptors.
Finally,
the query image is represented by the histogram of its visual words,
and similar reference images can be found as the ones with most similar histograms.
StreetMap then uses only the retrieved most similar reference images
for the subsequent framework steps of feature matching and RANSAC-based pose estimation.

Zhang~\etal~\cite{zhang2019high} proposed Micro-GPS.
They extract SIFT~\cite{lowe2004distinctive} features,
and map their descriptors to 8- or 16-dimensional vectors using principle component analysis.
An approximate nearest neighbour (ANN) search structure is constructed that incorporates $50$ randomly sampled features per reference image.
During localization,
this search structure efficiently matches query image features with their ANN among the reference image features.
Subsequently, 
a voting approach is adopted where each of the retrieved matches casts a vote for the camera position on a grid-divided map.
Only the matches voting for the grid cell that received most votes are used for RANSAC based pose estimation,
while others are rejected.

Schmid~\etal~\cite{schmid2020ground} build upon Micro-GPS~\cite{zhang2019high}.
They identified a drawback of Micro-GPS in the use of the ANN search structure. 
While it allows for efficient descriptor matching for initial localization,
it is not efficient for subsequent localization with available prior knowledge about the current robot pose.
This is because the use 
of an ANN search structure requires to match query image features with features of all reference images simultaneously.
For this,
Schmid~\etal substitute the use of ANN feature matching with the identity feature matching technique where only features with identical binary descriptor values are considered as matches.
This approach,
while still being efficient to compute,
allows to consider only a subset of reference images for feature matching,
e.g. the closest ones to the current pose estimate.
Subsequent steps of the localization pipeline are similar to that of Micro-GPS.

\section{Method}
We propose a deep learning framework for the retrieval of overlapping ground images.

Our goal is to solve the following problem:
Given a set of reference ground images $\sR$ and a query ground image $\vq \in \sQ$,
retrieve a set of similar reference images $\hat{\sR}_o \subset \sR$ to $\vq$ that should include all images $\sR_o \subset \sR$ that have overlapping content with $\vq$,
i.e. $\sR_o \subset \hat{\sR}_o \subset \sR$.

\subsection{Objective function}
Given two images $\vq \in \sQ$ and $\vr \in \sR$,
we normalize them 
and compute their embeddings $\ve_\vq$ and $\ve_{\vr}$.
The distance between $\vq$ and $\vr$ is computed with the L2-norm:
$\vd(\vq,\vr)=\|\ve_\vq-\ve_{\vr}\|_2$.
The actual overlap between the images is represented as $\vo(\vq,\vr)$,
it is computed as the proportion of the physical space that is covered by both images.
The goal of our training procedure is to adapt the weights of our CNN in such a way that $\vd(\vq,\vr) = 1 - \vo(\vq,\vr)$,
e.g. we want to have $\vd(\vq,\vr) = 1.0$ in the case of no overlap between $\vq$ and $\vr$,
$\vd(\vq,\vr) = 0.0$ in the case of full overlap.
For this purpose, 
we adopt the overlap loss of S{\'a}nchez-Belenguer~\etal~\cite{sanchez2020rise}:
\begin{equation}
\label{eq:2}
    \mathrm{L} =[\vd(\vq,\vr) - (1-\vo(\vq,\vr))]^2.
\end{equation}
But,
in contrast to~\cite{sanchez2020rise},
we employ 2D image overlaps,
which are available as the ground truth poses and the sizes of the areas covered by the images are known for training images.

\subsection{The model and its application}
\label{network}
Our network architecture and its training procedure is illustrated in Figure~\ref{fig:sn_nn}.
It consists of a CNN that extracts image features.
The activations of the final fully-connected layer represent the image embeddings.
We examine two variants of our architecture,
one using ResNet-50~\cite{he2016deep} as CNN backbone,
and the other using DenseNet-161~\cite{huang2017densely}.
We fix the unit and block configurations of both the DenseNets and ResNets as specified.
For convenience,
the proposed ground texture Deep Metric Learning (DML) models,
using DenseNet and ResNet backbones are henceforth referred to as DML-D and DML-R respectively.

During training,
input image pairs are processed in Siamese configuration~\cite{bromley1993signature, chopra2005learning},
which means that they are processed by identical CNNs with shared weights.
The network is trained with positive samples of actually overlapping pairs,
$\vq_1$ with $\vp_1$ and $\vq_2$ with $\vp_2$ in Figure~\ref{fig:sn_nn},
and negative samples of non-overlapping pairs,
$\vq_3$ with $\vn_1$ and $\vq_4$ with $\vn_2$ in Figure~\ref{fig:sn_nn}).
For each training sample,
the loss is computed according to Equation~(\ref{eq:2}) and backpropagated. 

For the image retrieval system,
depicted on the right of Figure~\ref{fig:sn_nn},
a k-d tree is built from all reference image embeddings.
Then,
at inference time,
it can be used to retrieve the reference images with most similar embeddings to that of the query image.

\section{Dataset and data preparation}
We employ the Micro-GPS ground image database~\cite{zhang2019high}.
The images are captured with a downward-facing PointGrey monochrome camera,
where each image covers an area of $0.2$m $\times$ $0.15$m ($1288$ $\times$ $964$ pixels).
The database contains images of six ground textures (Figure~\ref{fig:textures}).
For each texture,
a set of mapped reference images is provided.
The overall coverage area varies from about $12.75$m² to $41.76$m²,
which corresponds to $2014$ to $4043$ slightly overlapping reference images.
Additionally,
the database provides sequences of query images with different conditions and disruptions of the ground surface that were captured on different days,
at night with LED, with occlusion (e.g. leaves, dirt, and water),
and by lowering the shutter speed to increase motion blur.
We separate a sequence of $500$ query images per texture for the evaluation,
and use the other for parameter optimization and network training.

In order to prepare the data for training of our proposed DML-D and DML-R models,
we compute the pairwise overlap of each query image with all reference images.
This allows us to identify the positive training samples of query-reference image pairs for which we require to have at least $20\%$ overlap,
because we observed that the models can get confused by low-overlapping positive samples that become very similar to negative samples.
We train our models simultaneously on all textures,
because we aim for generalized models.
However,
the number of available positive samples varies for the different textures:
roughly $3000$ for concrete, tiles, and wood,
and roughly $10000$ for carpet, coarse, and asphalt.
To create additional training samples,
we apply random image augmentations of flips and rotations between $\pm45$ degrees.
Finally,
we obtain about 185,000 positive samples,
and the same number of negative non-overlapping samples is prepared.
These pairs are shuffled to be processed in random order to avoid processing multiple similar inputs in a row.

\section{Performance metrics}
\label{kpi}
We use recall as image retrieval performance metric.
Recall describes the share of correctly retrieved images.
It depends on the overall number of retrieved images $\vk=|\hat{\sR}_o|$,
i.e. the $\vk$ reference images with most similar descriptors to that of the query image,
and the maximum number of actually available correct retrievals among them.
So,
as a function of $\vk$,
we define:
\begin{equation}
\label{eq:3}
    \RatK = \frac{\mathrm{\#\,correctly\,\, retrieved\,\, images}}{\mathrm{\#\,maximum\,\, number\,\, of\,\, actually\,\, available\,\, correct\,\, retrievals}}.
\end{equation}
Whether a retrieved reference image is considered to be correct depends on its overlap with the query image.
Generally,
we are interested in all reference images with any overlap,
but the ones with large overlap are the most valuable ones for the localization task,
as they contain potentially the most correspondences of local image features with the query image.
This is why we compute $\RatK$ for varying overlap thresholds.
$\RiatK{\vx}$ represents the share of correctly retrieved reference images that have at least $\vx$ overlap with the query image.

Finally,
we employ the image retrievals for ground texture based initial localization.
Here,
we evaluate the localization success rate of Zhang~\etal\cite{zhang2019high},
where localization attempts are considered to be correct if the estimated query image pose has a translation difference of less than $4.8$\,mm
and an absolute orientation difference of less than $1.5$\,degrees.

\section{Evaluation}
\label{sec:eval}
Implementation details of the evaluated methods are presented in Section~\ref{sec:eval:impl},
this includes a survey about the optimal choice of the detector-descriptor pairing and the number of extracted local features per image for the BoW approach.
We examine image retrieval performance in Section~\ref{sec:eval:retrieval} 
and performance on the task of initial localization in Section~\ref{sec:eval:loc}. 
In all cases,
the number of retrieved most similar reference images is fixed to $\vk=100$.

In addition to our DML methods,
we evaluate BoW as the current state-of-the-art approach,
and we consider the deep metric metric learning approaches TL-MAC and AP-GeM.
These methods are trained with a classification loss,
which prevents us from training them for ground image retrieval.
However,
they have been found to have good generalization capabilities~\cite{Pion_Benchmark},
as they outperform other methods for visual localization tasks without being trained on the evaluation dataset.
Accordingly,
we employ these methods using pre-trained weights\footnote{https://github.com/naver/deep-image-retrieval}.
For TL-MAC the model was trained on the Landmarks-clean dataset~\cite{landmarks_dataset}.
For AP-GeM,
we achieve the best recall using weights of an instance
that was trained on the Google-Landmarks Dataset~\cite{GoogleLandmarks},
which has more than one million images from $15000$ places.
%
Furthermore,
we examine two baseline approaches.
The first is \textit{Random},
sampling $\vk=100$ reference images as retrieval result.
Our second baseline is to use the ResNet-50~\cite{he2016deep} and DenseNet-161~\cite{huang2017densely} CNNs
without task-specific fine tuning,
i.e. they are only pre-trained on ImageNet~\cite{Russakovsky_ImageNet}.

\subsection{Implementation}
\label{sec:eval:impl}
We implement our deep metric image retrieval approach in PyTorch\footnote{https://github.com/pytorch/pytorch/releases},
based on the Siamese network approach for image similarity with deep ranking of Wang~\etal~\cite{wang2014learning}.
Our ResNet-50~\cite{he2016deep} and DenseNet-161~\cite{huang2017densely} backbones are pre-trained on ImageNet~\cite{Russakovsky_ImageNet} for the task of object classification.
Subsequently,
we train them in Siamese configuration for ground image retrieval.
For more compact image embeddings,
we examined the replacement of the final pooling layers with generalized mean pooling layers~\cite{radenovic2018fine, cao2020unifying}.
However,
this decreased performance.
So,
we maintain the adaptive average pooling layers.
Also,
we experiment with node sizes of the final layer,
which defines the embedding size,
of $1000$, $2048$, and $4096$.
Best performance is reached with a size of $1000$.
During training,
we employ a batch size of 64 and tune the network weights using Adam optimizer with a learning rate of 10\textsuperscript{-4},
a weight decay of 10\textsuperscript{-5}.
After training,
we retrieve the $\vk=100$ reference images with most similar embeddings to the query image,
using the scikit-learn k-d tree\footnote{https://scikit-learn.org/stable/modules/neighbors.html}.
Our CNNs are trained and evaluated on 
five Titan X Pascal 12GB GPUs and an Intel Xeon E5-2630 v4 CPU at 2.20GHz. 

\begin{figure}[t]\TopFloatBoxes
    \begin{floatrow}
        \capbtabbox[\FBwidth]{%
          {\scriptsize
                \setlength{\tabcolsep}{2.5pt} 
                \begin{tabular}{|c|c|r|}
                    \hline
                    Detector&Descriptor&$\RatKi{100(\%)}$\\\hline\hline
                    SIFT     & SIFT       & \textbf{35.9}\\\hline
                    SURF     & SURF       & 9.4\\\hline
                    AKAZE    & AKAZE      & 17.4 \\\hline
                    SIFT     & LATCH      & 14.1\\\hline
                    AKAZE    & BRIEF      & 12.1\\\hline
                    AKAZE    & LATCH      & 13.7\\\hline
                \end{tabular}
            }
        }{%
            \caption{BoW $\RatKi{100}$ results on carpet for varying detector-descriptor pairings.}
            \label{tab:bow:pairings}
        }
        \capbtabbox[\FBwidth]{%
          {\scriptsize
                \setlength{\tabcolsep}{3.5pt} 
                \begin{tabular}{|l|rrrrrrrrrr|}
                    \hline
                    \multirow{2}{*}{Texture}&\multicolumn{10}{c|}{Number of features $-$ $\RatKi{100}$(\%)}\\\cline{2-11}
                    & 100 & 200 & 300 & 400 & 500 & 600 & 700 & 800 & 900 & 1000  \\\hline\hline
                    Carpet & 20.5 & 22.9 & 19.1 & 19.8 & 22.6 & 23.5 & 18.5 & \textbf{29.0} & 17.9 & 17.2 \\\hline
                    Asphalt (C) & 27.1 & 24.6 & 27.4 & 20.4 & 22.6 & \textbf{28.8} & 17.4 & 24.1 & 24.3 & 24.5 \\\hline
                    Concrete & 12.2 & 7.8 & 17.1 & 12.9 & 8.7 & 20.1 & 5.7 & \textbf{20.9} & 10.7 & 13.5 \\\hline
                    Asphalt (F) & \textbf{25.7} & 19.9 & 22.5 & 12.4 & 20.5 & 18.7 & 11.9 & 15.6 & 23.5 & 14.2\\\hline
                    Tiles & 8.9 & 18.1 & 17.7 & 12.8 & 11.1 & 14.1 & 17.0 & \textbf{20.5} & 16.6 & 16.0\\\hline
                    Wood & 6.1 & 9.4 & 9.2 & \textbf{11.4} & 5.7 & 11.3 & 8.6 & 8.3 & 7.9 & 9.4    \\\hline
                \end{tabular}
            }
        }{%
            \caption{BoW $\RatKi{100}$ results for all textures using varying numbers of extracted features per image.}
            \label{tab:bow:nof_features}
        }
    \end{floatrow}
\end{figure}

We implement the BoW approach,
using the FBOW library\footnote{https://github.com/rmsalinas/fbow} to create the vocabulary,
and we employ the OpenCV~4.0~\cite{bradski2000opencv} library to extract the required local visual features.
Here,
an important hyper-parameter choice is the type of employed local visual features.
We examine the keypoint detectors and feature descriptors that have been found to be the most successful ones for ground images according to Schmid~\etal\cite{schmid2019survey}:
SIFT~\cite{lowe2004distinctive}, SURF~\cite{Bay_SURF},
and AKAZE~\cite{alcantarilla2013AKAZE} are employed both as detectors and descriptors,
and additionally we examine the binary descriptors LATCH~\cite{levi2016latch} and BRIEF~\cite{calonder2010brief}.
The methods are parametrized with the optimized parameter settings provided by Schmid~\etal\cite{schmid2019survey}.
For the vocabulary creation,
a large set of features is required from the application domain.
Here,
we choose to extract $1000$ features from $1000$ reference images per texture.
In order to limit the number of extracted features per image,
we choose the features with largest keypoint response values
that have been assigned by the respective keypoint detectors.
Subsequently,
the vocabulary is used to map images to BoW representations
using their respective set of extracted features.
Here,
another important hyper-parameter choice is the size of this feature set $\vn$.
In the following,
we examine the optimal choice for both hyper-parameter settings.

First,
we set $\vn=1000$ and vary the detector-descriptor pairings.
We evaluate on the carpet texture and present results of $\RatKi{100}$ in Table~\ref{tab:bow:pairings}.
The combination of SIFT detector and SIFT descriptor clearly outperforms the other options.
Hence,
we use this variant in the following.

We also investigate the texture-specific optimal choice of $\vn$,
for values between $100$ and $1000$ with a step size of $100$.
Results of $\RatKi{100}$ are presented in Table~\ref{tab:bow:nof_features}.
In the following,
to achieve optimal BoW retrieval performance,
we always evaluate BoW image retrieval by selecting the respective texture-specific optimal choice of $\vn$.

\subsection{Evaluation of image retrieval performance}
\label{sec:eval:retrieval}
\begin{figure}[t]\TopFloatBoxes
    \begin{floatrow}
        \ffigbox[\FBwidth]{%
          \includegraphics[width=0.56\textwidth]{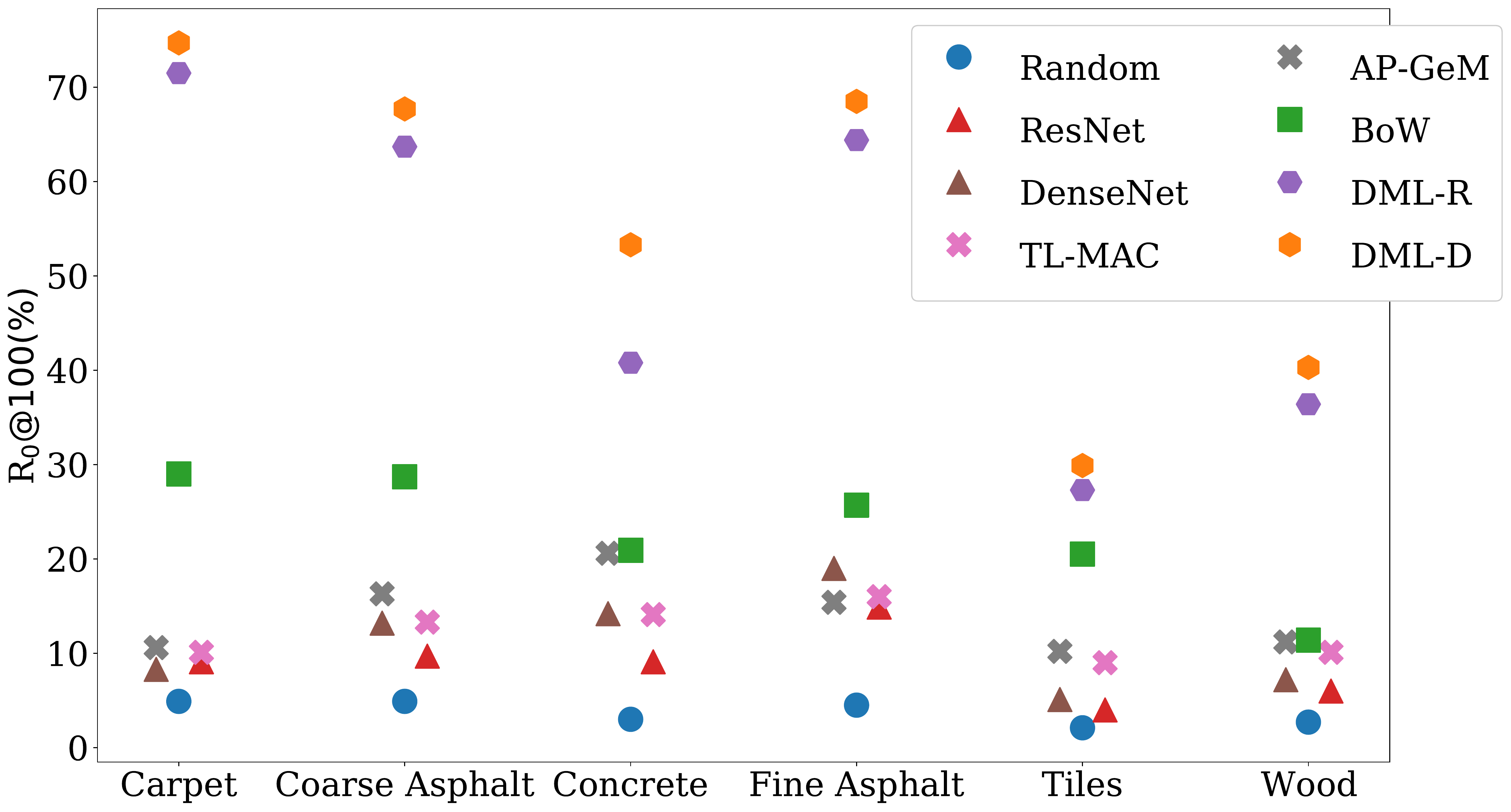}
        }{%
          \caption{$\RiatKj{0}{100}$ results for all textures.}
          \label{fig:plot_fig}
        }
        \hspace{-0.3cm}
        \capbtabbox[\FBwidth]{%
          \setlength{\tabcolsep}{3.5pt} 
          {\scriptsize
    		\begin{tabular}{|l|r|r|r|r|r|}
    			\hline
    			\multirow{2}{*}{Model} & \multicolumn{5}{c|}{$\RatKi{100}$(\%)} \\\cline{2-6}
    			& \multicolumn{1}{|c}{R\textsubscript{0}}& \multicolumn{1}{|c}{R\textsubscript{20}}& \multicolumn{1}{|c}{R\textsubscript{40}}& \multicolumn{1}{|c}{R\textsubscript{60}}& \multicolumn{1}{|c|}{R\textsubscript{80}}\\\hline\hline 
    			Random & 3.7 & 3.7 & 3.7 & 3.8 & 4.0\\\hline
    			ResNet &	8.8	& 11.3 & 14.0 & 17.1 & 25.2\\\hline
    			DenseNet & 11.2 & 15.9 & 20.8	& 25.5 & 41.5\\\hline
    			TL-MAC & 12.2 & 17.4 & 22.4 & 28.7 & 41.3 \\\hline
    			AP-GeM & 14.1   & 18.9 & 23.6 & 29.0 & 39.0 \\\hline
    			BoW &  22.7& 42.3& 64.4 & 82.9 & 93.5\\\hline
    			DML-R &	51.0&	68.0&	81.9	& 90.5 & 93.5\\\hline
    			\textbf{DML-D} & \textbf{55.7}	& \textbf{75.0}	& \textbf{89.5}& \textbf{97.0}	& \textbf{99.3}				\\ \hline
    		  \end{tabular}
    		}
        }{%
          \caption{
    		$\RatKi{100}$ results averaged over all textures for varying overlap thresholds.
    		The respective best recall results are highlighted in \textbf{bold}.}
    	  \label{tab:recall}
        }
    \end{floatrow}
\end{figure}

Figure~\ref{fig:plot_fig} presents the results for $\RiatKj{0}{100}$.
The random baseline has the lowest recall results.
Varying performance of this method for the different textures
can be explained by the correspondingly varying number of reference images.
Our models have the best retrieval performance,
clearly outperforming BoW,
the current state-of-the-art for ground image retrieval.
DML-D, with the DenseNet-161 backbone,
achieves slightly better performance than DML-R, 
using a ResNet-50 backbone.
Also,
we observe that fine-tuning the models for the use on ground images is of great importance,
as our DML methods perform much better than the networks
that are have not been trained on ground images:
ResNet, DenseNet, TL-MAC, and AP-GeM.
This can be explained by the fact
that the random patterns observed in ground images are quite different
to the structured environments of, for example, ImageNet.

Matching our results,
Zhang~\etal~\cite{zhang2019high} and Schmid~\etal\cite{schmid2020ground}
identified concrete and wood to be the most challenging of the evaluated textures,
but they had good results on tiles.
However,
concrete, wood, and tiles are also the textures
for which we have only about $3000$ samples of overlapping query-reference image pairs
(without synthetic augmentation),
while we have $10000$ for the others.
This might be the main reason for our poor performance on tiles
and it adds to the challenge on concrete and wood.

\begin{figure}[t]\TopFloatBoxes
    \begin{floatrow}
        \ffigbox[\FBwidth]{%
          \includegraphics[width=0.465\textwidth]{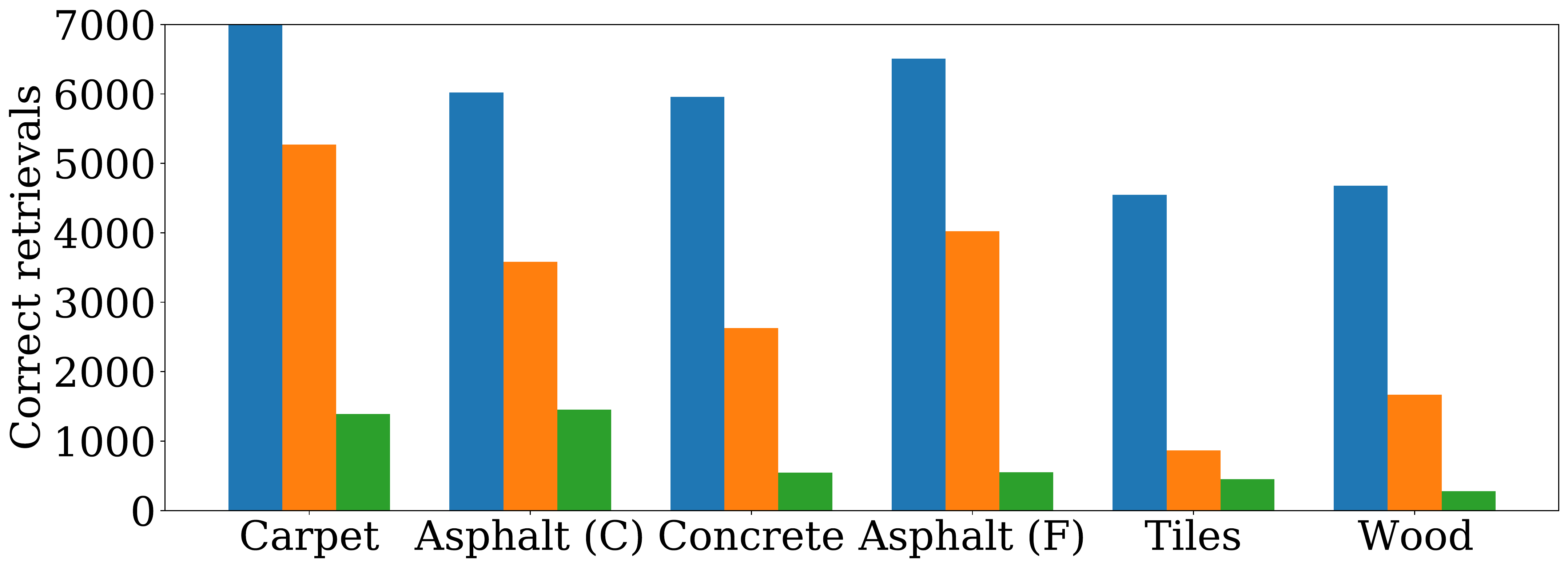} 
        }{%
          \vspace{-0.15cm}
          \caption{Number of correct retrievals, and\\ actually available ones, with $<40\%$ overlap.}
          \label{fig:less_than_40}
        }
        \hspace{-0.2cm}	
        \ffigbox[\FBwidth]{%
          \includegraphics[width=0.465\textwidth]{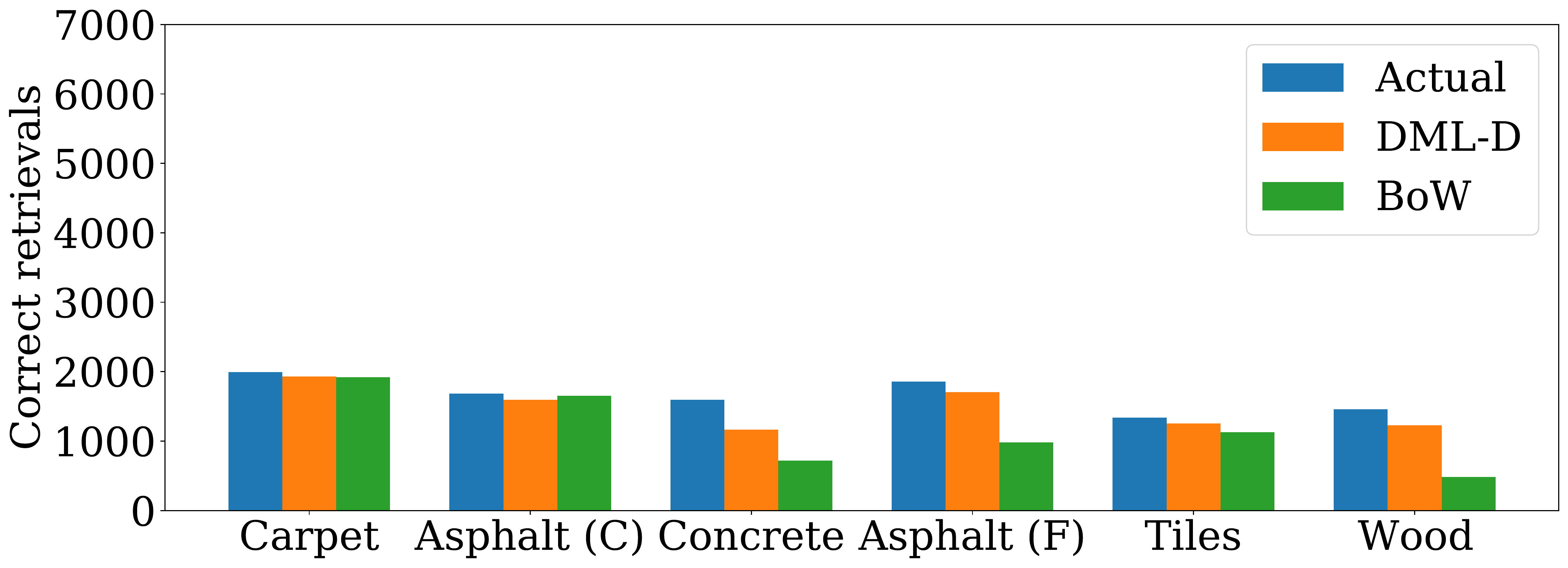}
        }{%
            \vspace{-0.15cm}
            \caption{Number of correct retrievals, and actually available ones, with $\ge40\%$ overlap.}
            \label{fig:at_least_40}
        }
    \end{floatrow}
\end{figure}

\begin{figure}[t]
	\includegraphics[width=0.241\textwidth]{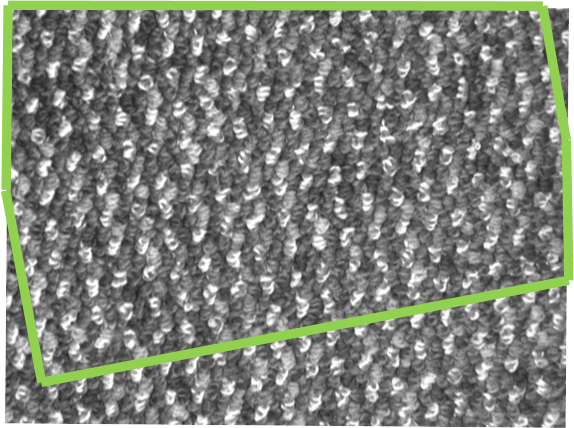}
	\includegraphics[width=0.241\textwidth]{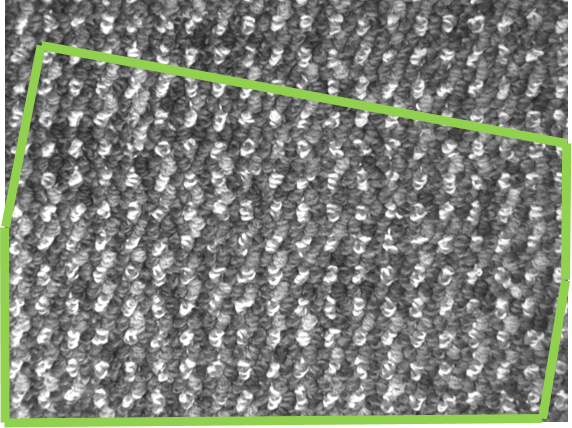}\hspace{2.1mm}
	\includegraphics[width=0.241\textwidth]{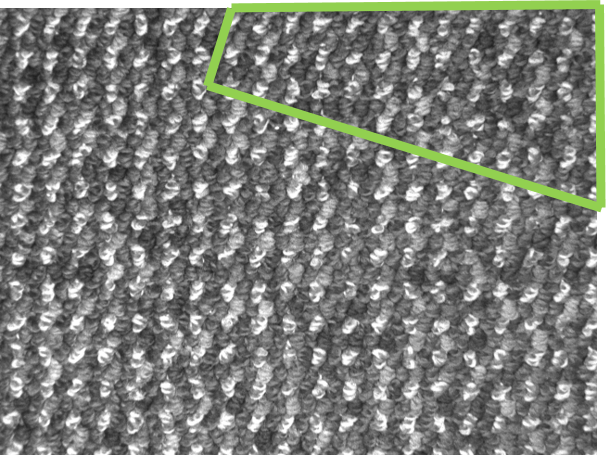}
	\includegraphics[width=0.241\textwidth]{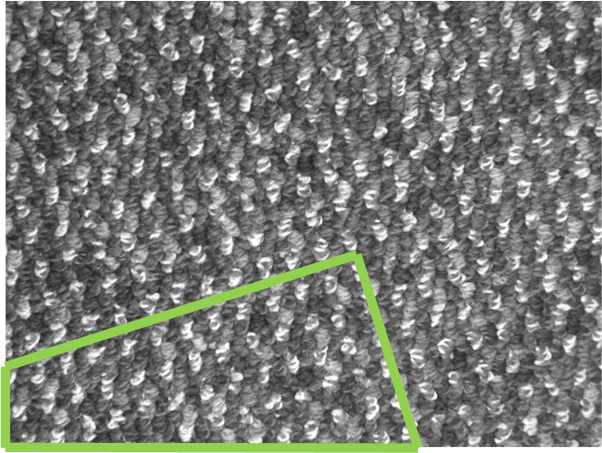}
   \vspace{-0.60cm}	
   \caption{
            Two examples from the carpet texture,
            where the top DML-D retrievals are correct with $75.8\%$ and $19.9\%$ overlap.
            The green border indicates the overlapping area.}
\label{fig:correct_retrievals}
\end{figure}

Table~\ref{tab:recall} presents $\RatKi{100}$ averaged over all textures.
Different thresholds for the minimum required overlap of the retrievals are considered,
i.e. [R\textsubscript{0},..,R\textsubscript{80}]@100.
Our DML-D model has the best retrieval performance with an average $\RiatKj{\vx}{100}$ of $83.3\%$,
outperforming the BoW approach with an average of $61.2\%$.
The reference images with large overlap are correctly retrieved by the BoW approach.
Hence,
it achieves a $\RiatKj{80}{100}$ of $93.5\%$.
However,
most of the reference images with only small amounts of overlap with the query image are not correctly retrieved,
which leads to poor recall values for $\RiatKj{0}{100}$ and $\RiatKj{20}{100}$ of only $22.7\%$,
respectively $42.3\%$.
We investigate this further by comparing the numbers of correctly retrieved reference images with less than $40\%$ overlap with the query image (Figure~\ref{fig:less_than_40}),
and with at least $40\%$ overlap (Figure~\ref{fig:at_least_40}).
Figure~\ref{fig:correct_retrievals}
presents examples where DML-D correctly retrieved images with more,
respectively less,
than $40\%$ overlap.
As expected,
we observe BoW to be competitive with DML-D for retrieving the images with large overlaps,
while it gets outperformed by DML-D for images with small overlaps.
This indicates a better representation of our learned image embeddings compared to the BoW image representations.

It is also of interest to examine how often image retrieval failed completely,
i.e. not a single overlapping reference image is retrieved,
because in these cases subsequent successful pose estimation based on the retrieved images is impossible. 
For DML-D,
we observe a total of $18$ failure cases,
$17$ on concrete and one on wood.
The BoW approach has $201$ failure cases,
$135$ on wood, $53$ on concrete, $7$ on tiles,
$5$ on fine asphalt, and one on coarse asphalt.

\subsection{Evaluation of initial localization success rate}
\label{sec:eval:loc}
Finally,
we evaluate the localization success rate of initial localization,
using BoW and DML-D image retrievals.
Here,
we employ the localization method of Schmid~\etal\cite{schmid2020ground},
for which the implementation was provided to us by the authors.
The first input of this method is the query image,
and the second input is the set of reference images to be considered for potential feature correspondences.
We examine the localization success rate (a) using all available references images,
(b) using the top-100 image retrievals of the BoW approach or (c) that of our DML-D model. 
On average over our sequences of $500$ query images,
the localization method has a success rate of $95.5\%$ without image retrieval,
$87.3\%$ when using BoW image retrieval and $96.6\%$ with DML-D image retrievals.
For comparison,
the average success rate with AP-GeM is $49.9\%$.
The success rate without retrieval and with DML-D retrievals is close to $100\%$ for carpet,
tiles, coarse, and fine asphalt.
On concrete,
it is slightly decreased using DML-D ($92.8\%$) compared to the case without image retrieval ($97.8\%$),
while it is significantly increased on wood,
the most challenging texture,
with $87.0\%$ with DML-D to $75.0\%$ without retrieval (and $45.6\%$ with the BoW retrievals).
Generally,
the application of image retrieval also has the advantage of reducing the required computation time for localization.
According to Schmid~\etal~\cite{schmid2020ground},
using just $100$ instead of all $2014$ references images of the carpet texture reduces the computation time for feature matching from $286.47$\,ms to only $15.25$\,ms.
A robotic agent using our image retrieval method could therefore localize faster,
which can be highly beneficial in practice. 

\section{Conclusion}
We introduced a deep learning approach to image retrieval of ground images,
using a CNN trained in Siamese fashion for the task of predicting the overlap of image pairs.
Our method significantly outperforms Bag-of-Words (BoW),
representing the current state of the art for the task.
Also,
the image retrievals of our method are significantly better suited for initial localization than that of the BoW approach.

In this work,
we examined generalized models,
being trained on all textures simultaneously.
We also examined the performance of our models if trained texture-specifically,
but this did not clearly improve the image retrieval recall.
The resulting localization success rate was even slightly lower ($96.0\%$ to $96.6\%$).
For future research,
we would like to investigate the generalization performance to textures not being included in the training process.
Also,
for larger application areas than that of the here considered database with a maximum of $41.76$m²,
with up to $4043$ reference images,
we expect to observe even larger advantages in the localization task when using our image retrievals compared to a setup without image retrieval.
However,
this remains to be tested on upcoming ground image databases. 

\bibliography{./bib}
\end{document}